\documentclass[fleqn,10pt]{wlscirep}

\usepackage[utf8]{inputenc}
\usepackage[T1]{fontenc}
\usepackage{array}
\usepackage{multirow}
\usepackage{subcaption}

\title{FZ-VLM: A Two Stage Florence-Zephyr Vision Language Model Framework for Pulmonary Nodule Characterization and Clinical Decision Making}

\author[1,*]{Pramit Dutta}
\author[2]{Jenita Manokaran}
\author[3]{Richa Mittal}
\author[4]{Ryan Appleby}
\author[1]{Eranga Ukwatta}

\affil[1]{College of Engineering, University of Guelph, Guelph, ON N1G 2W1, Canada}
\affil[2]{Holland Bloorview Kids Rehabilitation Hospital, Toronto, ON M4G 1R8, Canada}
\affil[3]{Guelph General Hospital, Guelph, ON N1E 4J4, Canada}
\affil[4]{Department of Clinical Studies, Ontario Veterinary College, University of Guelph, Guelph, ON N1G 2W1, Canada}

\affil[*]{Corresponding author: eukwatta@uoguelph.ca}


\begin{abstract}

Lung cancer remains one of the leading causes of cancer related mortality worldwide, and Computed Tomography (CT) is a primary imaging tool for screening and followup assessment. After pulmonary nodule detection, radiologists manually assess anatomical location, diameter, margin characteristics, and attenuation type to support risk assessment and clinical decision making. However, this post detection workflow is time consuming and can be affected by inter observer variability. Existing Artificial Intelligence methods often focus on isolated tasks, which limits their use as a unified and clinically grounded interpretation framework. This study presents FZ-VLM, a two stage Florence-Zephyr Vision Language Model framework for unified structured pulmonary nodule characterization in lung CT. The framework uses a fine-tuned Florence-2 model to extract radiological attributes from expert annotated 2D axial CT slices, while a Zephyr-7B model uses the extracted attributes to generate nodule descriptions, follow-up recommendations, and longitudinal analysis. A structured Visual Question Answering dataset containing 8,540 image-prompt-answer triplets was curated from 745 patient cases derived from the National Lung Screening Trial. Results showed that the Stage 1 model achieved 77.18\% accuracy for anatomical location, 67.96\% accuracy for margin characteristics, and 79.13\% accuracy for attenuation type, with a Mean Absolute Error of 2.58 mm for diameter estimation, outperforming evaluated GPT-4-based baselines as well as human baseline. Expert radiologist evaluation of Stage 2 showed 93.9\% accuracy, 98.6\% completeness score, 76.1\% clinical relevance, and an overall score of 89.5\%. Safety analysis showed that most outputs were clinically safe, although some follow up recommendations still required expert review. To the best of our knowledge, this study presents the first two stage Vision Language Model framework for structured nodule characterization and clinical decision making system.

\end{abstract}

\begin{document}

\flushbottom
\maketitle

\thispagestyle{empty}

\section*{Introduction}

Lung cancer is the leading cause of cancer-related mortality worldwide, causing nearly 1.8 million deaths each year \cite{Sung2021}. Early diagnosis can increase the five-year survival rate to approximately 65.5\%, compared with less than 29.5\% overall survival \cite{SEER2026}. Low dose computed tomography (LDCT) is the primary imaging modality for lung cancer screening and follow-up assessment \cite{NLST2011}.

The clinical workflow considered in this study consists of three stages. The process begins with high-risk lung cancer screening and LDCT image acquisition, followed by CT volume review for nodule identification. After a pulmonary nodule is detected, the radiologist selects the representative CT slice where the longest nodule diameter is most visible. In the post-detection stage, key radiological attributes, including anatomical location, diameter, margin characteristics, and attenuation type, are extracted to support risk assessment, follow-up recommendations, and treatment planning \cite{Gould2013,Dutta2026}.

The proposed framework focuses on this post-detection interpretation stage, where structured nodule characterization is required for clinical decision making \cite{ACR2022}. In this study, four parameters were considered: anatomical location, diameter, margin characteristics, and attenuation type. Anatomical location identifies the affected lung lobe \cite{Liu2017}, while in the curated NLST annotations nodule diameter is calculated as the average of the long-axis and perpendicular short-axis measurements, rounded to the nearest whole millimeter, and used to support size-based management \cite{ACR2022}. Margin characteristics describe the nodule boundary \cite{Ferreira2018,Xie2019}, whereas attenuation represents its internal density pattern \cite{Godoy2009,Zhou2020}. Accurate extraction of these attributes is therefore important for risk stratification and follow-up decisions.

However, manual attribute extraction is time consuming and affected by interobserver variability, particularly for qualitative features such as margin and attenuation. Unanimous correct classification among expert radiologists has been reported in only 58\% of pulmonary nodule cases \cite{Han2018,Ridge2016}. To address this issue several deep learning methods have been developed for detection, segmentation, location assessment, diameter estimation, margin characterization, and attenuation classification
\cite{Manokaran2024,Setio2017,Ferreira2018,Zhou2020}. However, most methods remain task specific and do not provide a unified framework for structured nodule characterization \cite{Dutta2026}.

Vision Language Models (VLMs) offer a possible solution because they connect visual information with language prompts and structured questions. General-purpose VLMs support image-text alignment, visual question answering, grounding, and instruction following \cite{Radford2021,Liu2023,Xiao2023}. However, their direct use in radiology remains limited because CT findings involve subtle patterns, domain-specific labels, and clinically sensitive decisions \cite{NDutta2026,LLiu2024}. Medical adaptation through prompt learning, fine-tuning, structured supervision, and clinically grounded image evidence is therefore important \cite{NDutta2026,Veasey2024,Chen2025}.

Structured visual question answering is particularly suitable for pulmonary nodule analysis because it converts broad image interpretation into focused questions about location, size, margin, and attenuation \cite{ACR2022}. Recent studies suggest that structured VQA and guideline-based prompting can improve clinically meaningful interpretation \cite{Chang2025,Khademi2025,Shi2026,Zhao2026}. However, medical VLMs may still produce fluent responses that are weakly grounded in the image, creating risks of hallucination and unsafe recommendations \cite{Dutta2026,NDutta2026}. Reliable systems should therefore use clinically constrained outputs and task-specific evaluation. Large language models can provide this reasoning layer that organizes extracted findings and generates structured clinical explanations. Their use should remain assistive because overconfident or poorly calibrated outputs may cause harm \cite{Yang2023,McDuff2025,Hager2024}.

Explainability is also important for clinical use. Visual evidence can indicate whether a model focuses on the relevant lesion, while textual reasoning can show how extracted attributes support the generated interpretation. Prior studies have used saliency maps, attention visualization, quantitative explanation metrics, radiomics, and guideline-based reasoning to improve transparency \cite{Rao2023,Brima2024,Komorowski2023,Lin2024,Shi2026}. A clinically useful framework should therefore provide both image-level and language-level explanations.

Large screening datasets remain underused for structured VLM development. The National Lung Screening Trial (NLST) is a major multicenter and multivendor lung cancer screening dataset, but it was not designed for VQA \cite{NLST2011}. Recent lung nodule VLM studies more commonly use LIDC-IDRI because it provides accessible semantic annotations and segmentation labels \cite{Khademi2025,Shi2026,Zhao2026}. NLST therefore requires additional curation to create image-prompt-answer triplets for attribute-based learning.

These limitations indicate the need for a unified framework that can extract multiple pulmonary nodule attributes from CT images and use them as controlled inputs for clinical reasoning. Therefore, this study proposes FZ-VLM, a two-stage Florence-Zephyr framework for structured pulmonary nodule characterization and clinical decision support. The first stage uses a fine-tuned Florence-2 model to extract radiological attributes from expert-selected 2D axial CT slices \cite{Xiao2023}. The second stage provides these image-derived attributes to Zephyr-7B to generate nodule descriptions, follow-up recommendations, and longitudinal analyses \cite{Zephyr7b}. This attribute-to-reasoning design limits unrestricted language generation and helps ground the generated outputs in explicit radiological findings \cite{Liu2024}. The study also transforms NLST imaging data and associated annotations into structured image-prompt-answer triplets for model development and evaluation \cite{NLST2011}.

The framework was evaluated through quantitative analysis, image-level explainability for qualitative analysis, slice-sensitivity analysis, baseline comparison, external generalization assessment, and expert radiologist review. The main contribution of this work is a unified pipeline that connects visually grounded nodule characterization with structured language-based clinical interpretation. The study further contributes an NLST-derived VQA dataset and two complementary forms of interpretability: Florence-2 attention heatmaps for image-level evidence and Zephyr-7B reasoning for language-level explanation. To the best of our knowledge, this is the first two-stage VLM framework developed specifically for structured pulmonary nodule characterization and subsequent clinical decision support.

\section*{Methods}

This study developed a two stage Florence–Zephyr Vision Language Model (FZ-VLM) for structured interpretation of pulmonary nodules in low dose CT images. The methodological workflow comprised dataset curation, Florence-2 fine-tuning, hierarchical integration of the extracted attributes, zero-shot clinical interpretation using Zephyr-7B and overall evaluation strategy.

\subsection*{Data Curation and Dataset Construction}

Low dose chest CT examinations were obtained from the National Lung Screening Trial dataset. Cases were selected from screen detected and non-screen detected lung cancer cohorts for which confirmed disease status, nodule location, representative slice index, and expert annotated nodule attributes were available. Cases with missing, indeterminate, or incomplete annotations were excluded. 

For each nodule, the axial slice containing the maximum nodule diameter was selected using the reference slice index. CT intensities were converted from Hounsfield units to 8-bit image intensities using a lung window with a width of 1700 HU and a level of $-700$ HU. Values below $-1550$ HU and above 150 HU were clipped before linear intensity normalization. The selected window was applied uniformly to all images.

Then, each CT image was paired with standardized questions targeting four radiological attributes, and the attribute as answer. The resulting dataset contained 8,540 image-question-answer triplets, divided into training (7,332), validation (384), and held-out test (824) sets using an approximately 85:5:10 ratio. Splitting was performed at the patient level so that all images and attribute questions associated with an individual patient remained in the same partition, thereby preventing information leakage between training and evaluation data.

\subsection*{Florence–Zephyr Vision Language framework}

\begin{figure}[ht]
    \centering
    \includegraphics[width=\textwidth]{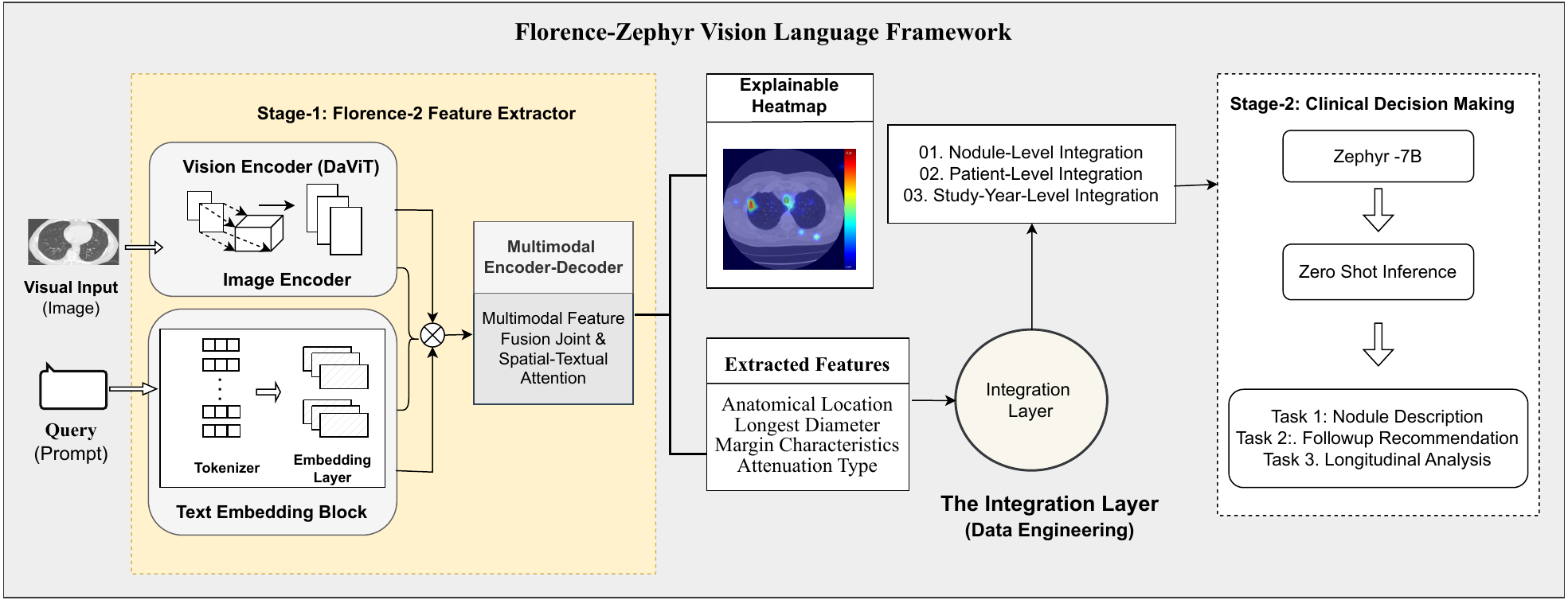}
    \caption{Architecture of the FZ-VLM framework. The pipeline executes attribute extraction by a finetuned Florence-2 with explainable attention heatmaps. These parameters are processed through a hierarchical integration layer to serve as contextual input for a Zephyr-7B model that performs zero-shot inference to generate structured clinical outputs.}
    \label{fig:fz_vlm}
\end{figure}

We developed the Florence–Zephyr Vision-Language Model, or FZ-VLM, as a two-stage framework for structured pulmonary nodule interpretation, as illustrated in Figure~\ref{fig:fz_vlm}. The first stage uses a domain-adapted Florence-2 model to extract visually grounded radiological attributes from CT images. The second stage uses Zephyr-7B to transform the extracted attributes into structured nodule descriptions, follow-up recommendations, and longitudinal interpretations.

The two-stage design separates image perception from clinical language generation. Consequently, the reasoning model receives an explicit set of radiological attributes rather than unrestricted visual embeddings. This design allows the generated clinical interpretation to be traced to specific model-extracted findings and permits the visual and reasoning components to be evaluated independently.

\subsection*{Stage 1: Florence-2 Feature Extractor}

In Stage 1 of the FZ-VLM framework the large variant of Florence-2 was used as the visual attribute extraction model. The model contains approximately 0.77 billion parameters and includes a DaViT vision encoder, a text embedding block, and a multimodal encoder-decoder architecture. Each input consisted of a windowed axial CT image paired with one of four standardized questions targeting anatomical location, diameter, margin characteristics, or attenuation type. The model generated the corresponding radiological attribute as a text response.

The pretrained Florence-2 model was fine-tuned using the training subset of the curated VQA dataset and evaluated during training using the validation subset. This task-specific adaptation enabled the model to learn the relationship between visual features in low-dose CT slices and the corresponding expert-annotated radiological attributes. Fine-tuning was conducted on the Rorqual computing cluster provided by the Digital Research Alliance of Canada using two NVIDIA H100 GPUs. The model was trained using the AdamW optimizer with a base learning rate of $2\times 10^{-6}$. A cosine decay learning rate scheduler with two warm up steps was applied over 10 training epochs. The batch size was two image-question-answer triplets per device. The trained model was subsequently evaluated on the held out unseen test set.

\subsection*{Hierarchical integration of extracted attributes}

\begin{figure}[ht]
    \centering
    \includegraphics[width=\textwidth]{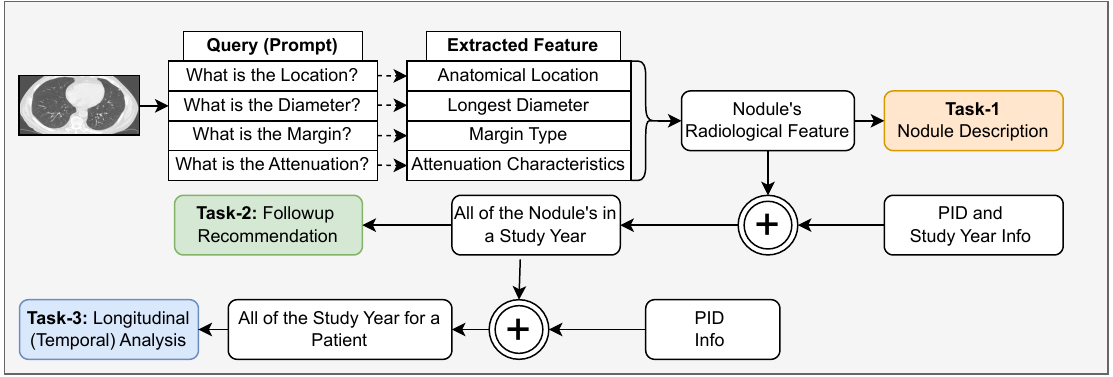}
    \caption{Hierarchical integration of the radiological attributes extracted by Florence-2. Individual nodule features are combined into various profiles for different assessment.}
    \label{fig:integration_layer}
\end{figure}

The four outputs generated by Florence-2 were combined through a hierarchical integration layer, as illustrated in Figure~\ref{fig:integration_layer}. At the nodule level, the extracted radiological attributes were assembled into one structured nodule representation for nodule description. At the study-year level, all nodules identified within a screening examination were grouped to represent the complete imaging findings and support follow-up recommendations. At the patient level, the study-year representations were ordered chronologically to create a longitudinal record for temporal analysis.

\subsection*{Stage 2: Clinical Decision Making using Zephyr-7B}

Zephyr-7B was used as the clinical decision making component. The model is an instruction aligned language model derived from the Mistral-7B architecture and optimized through direct preference optimization. In the proposed framework, Zephyr-7B was used without additional fine-tuning. Structured attributes generated by Stage 1 were inserted into standardized prompt templates. 

Three clinical tasks were evaluated: the nodule description task generated a concise radiological description from the attributes of an individual nodule. The follow-up recommendation task produced an overall recommendation based on all nodules identified during a single screening examination. The longitudinal analysis task interpreted interval changes across multiple screening examinations, including changes in nodule number, size, margin, and attenuation.

Task-specific system instructions defined the model as a medical assistant and directed it to base its response exclusively on the provided structured findings. The prompts did not include demographic, pathological, or outcome information that was unavailable from Stage 1. This constraint was used to reduce unsupported inference and maintain traceability between the generated output and the extracted radiological evidence.

\subsection*{Overview of Experimental Evaluation}

The proposed FZ-VLM framework was evaluated using unseen test data that was not used during model training or validation. The evaluation was designed to assess both stages of the framework separately. This two level evaluation strategy was used to examine whether the framework could first extract reliable radiological attributes from CT images and then use those attributes to generate clinically meaningful interpretation.

In Stage 1, the fine-tuned Florence-2 model was assessed through five evaluation phases: standard quantitative performance analysis, image level explainability assessment,  slice selection sensitivity analysis, baseline comparison, and external validation. Standard evaluation parameters were used to measure attribute extraction performance, including accuracy, precision, recall, F1-score, confusion matrix analysis, and mean absolute error for diameter estimation.

In Stage 2, the Zephyr-7B model was evaluated through expert radiologist review. The evaluation focused on generated clinical outputs, and each output was reviewed using clinically relevant parameters. Accuracy assessed whether the generated output correctly represented the clinical finding or recommendation. Completeness evaluated whether the output included the necessary clinical information without missing important details. Clinical relevance assessed whether the interpretation was meaningful for pulmonary nodule assessment. Safety evaluated whether the output avoided unsupported, misleading, or potentially harmful clinical recommendations.

\section*{Results}

This section presents the experimental evaluation of the proposed FZ-VLM framework. The evaluation is divided into two main stages, where Stage 1 is assessed for radiological attribute extraction and Stage 2 is evaluated for generated clinical interpretation via expert radiologist review. Additional analyses included explainability assessment, slice sensitivity testing, baseline comparison, and external generalization.

\subsection*{Quantitative Performance Analysis of Florence-2}

For radiological attribute extraction, the fine-tuned Florence-2 model achieved an overall accuracy of 74.42\% with 77.18\% for anatomical location, 67.96\% for margin characteristics, and 79.13\% for attenuation type on the held-out test set. Table~\ref{tab:florence_summary} summarizes the macro average precision, recall, and F1-score for the three categorical radiological attributes. Anatomical location extraction showed the strongest macro F1-score among the categorical tasks, while margin characteristics showed moderate performance. Attenuation type achieved the highest task-level accuracy, although its lower macro average scores indicate reduced performance for minority classes.

\begin{table}[ht]
    \centering
    \caption{Performance of the fine-tuned Florence-2 model on the test set. Macro average precision, recall, F1-score, and Accuracy are reported for categorical attributes. For  diameter prediction, accuracy was measured using a tolerance from 0 to 5 mm.}
    \label{tab:florence_summary}
    \small
    \begin{tabular}{|p{4.2cm}|>{\centering\arraybackslash}p{2.5cm}|>{\centering\arraybackslash}p{2.5cm}|>{\centering\arraybackslash}p{2.5cm}|>{\centering\arraybackslash}p{1.2cm}|}
    \hline
    Parameter & Precision & Recall & F1-score & Accuracy \\
    \hline
    Anatomical location & 0.75 & 0.72 & 0.70 & 77.18\% \\
    \cline{1-5}
    Margin characteristics & 0.65 & 0.64 & 0.64 & 67.96\% \\
    \cline{1-5}
    Attenuation type & 0.58 & 0.52 & 0.53 & 79.13\% \\
    \cline{1-5}
    Nodule diameter & \multicolumn{4}{|c|}{\parbox{10cm}{\centering MAE = 2.58 mm; SD = 3.73 mm; Accuracy = 21.4--87.9\%}} \\
    \hline
    \end{tabular}
\end{table}

For diameter prediction, the model achieved an MAE of 2.58 mm with an SD of 3.73 mm. The result also shows that accuracy increased from 21.4\% at exact match to 87.9\% within $\pm$5 mm tolerance. Although the model showed relatively low exact match accuracy, it performed reliably within small measurement tolerances. This suggests that the predicted diameters were generally close to the reference measurements, as also reflected by the median absolute error of 1 mm.

\subsection*{Explainability Analysis Using Attention Heatmaps}

To analyze the decision-making behavior of the fine-tuned Florence-2 model, attention heatmaps were generated for representative test cases. These heatmaps provide a visual indication of the image regions that received higher model attention during radiological attribute extraction. In this analysis, selected cases were examined to understand how the model focused on nodule-related regions, how correct predictions were formed, and how different types of prediction errors appeared across anatomical location, diameter, margin characteristics, and attenuation type.

\begin{figure*}[ht]
    \centering
    
    \begin{subfigure}{0.3\textwidth}
        \centering
        \includegraphics[width=\linewidth]{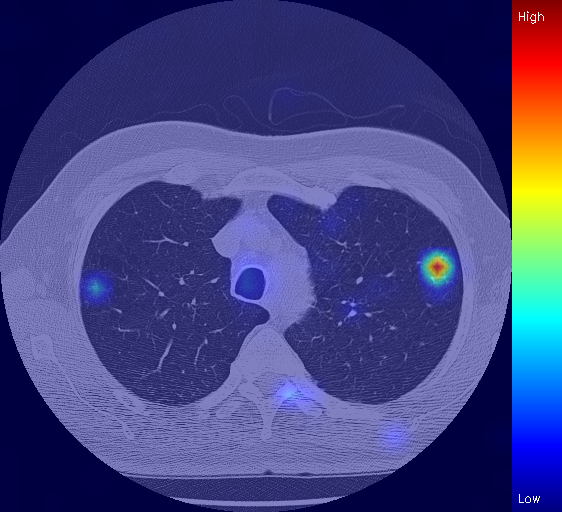}
        \caption{Case 1: LUL, 5 mm, poorly defined, ground glass. Model: LUL, 7 mm, poorly defined, ground glass.}
    \end{subfigure}
    \hfill
    \begin{subfigure}{0.3\textwidth}
        \centering
        \includegraphics[width=\linewidth]{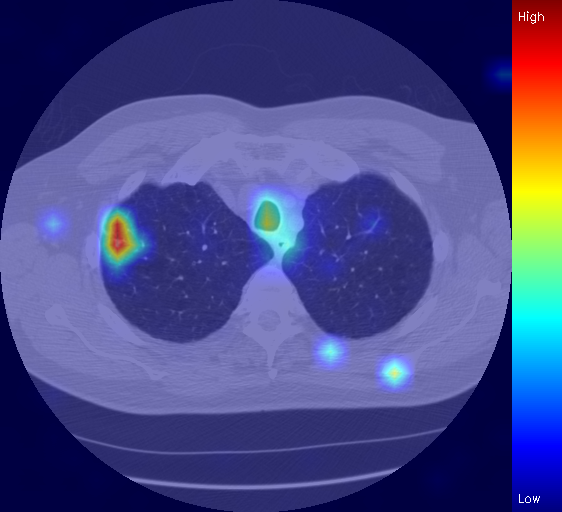}
        \caption{Case 2: RUL, 6 mm, poorly defined, soft tissue. Model fully matched the reference attributes.}
    \end{subfigure}
    \hfill
    \begin{subfigure}{0.3\textwidth}
        \centering
        \includegraphics[width=\linewidth]{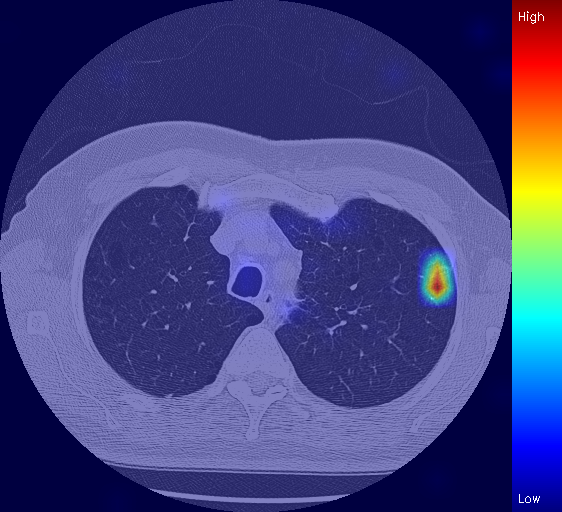}
        \caption{Case 3: LUL, 15 mm, poorly defined, ground glass. Model: LUL, 13 mm, poorly defined, ground glass.}
    \end{subfigure}
    
    \vspace{0.4cm}
    
    \begin{subfigure}{0.3\textwidth}
        \centering
        \includegraphics[width=\linewidth]{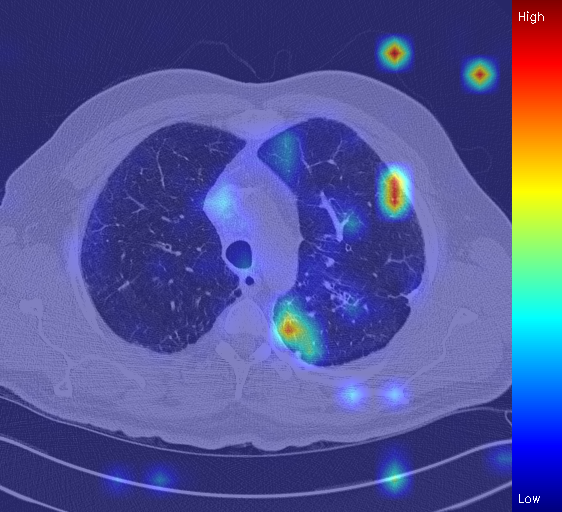}
        \caption{Case 4: LUL, 5 mm, smooth, soft tissue. Model fully matched the reference attributes.}
    \end{subfigure}
    \hfill
    \begin{subfigure}{0.3\textwidth}
        \centering
        \includegraphics[width=\linewidth]{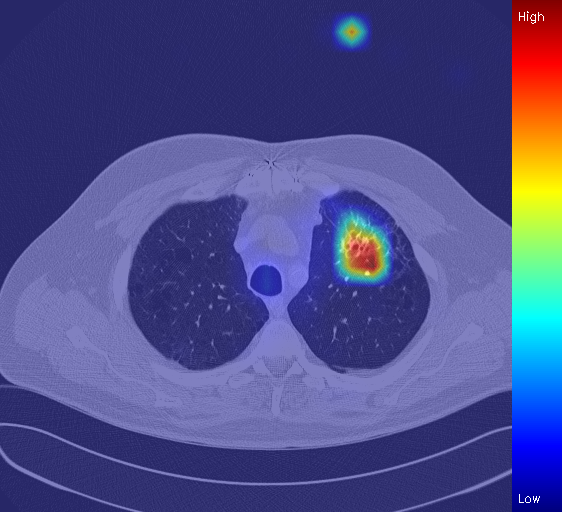}
        \caption{Case 5: LUL and margin were correct, but diameter and attenuation were incorrect.}
    \end{subfigure}
    \hfill
    \begin{subfigure}{0.3\textwidth}
        \centering
        \includegraphics[width=\linewidth]{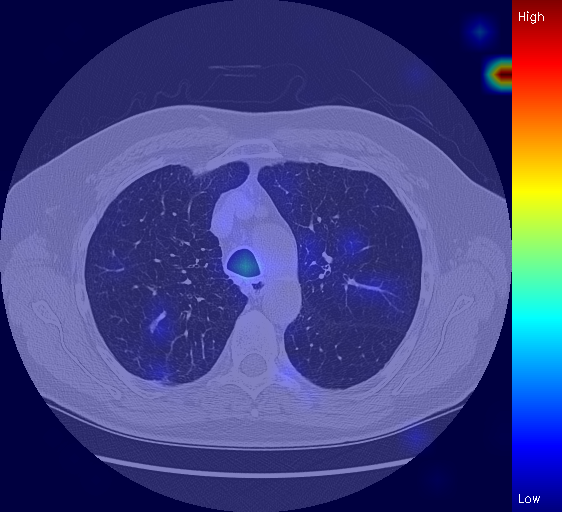}
        \caption{Case 6: Severe error case with incorrect location, diameter, margin, and attenuation.}
    \end{subfigure}
    
    \caption{Qualitative explainability and error analysis using attention heatmaps. Representative cases show correct predictions and error patterns in Stage 1 radiological attribute extraction. LUL: left upper lobe; RUL: right upper lobe.}
    \label{fig:qualitative_heatmaps}

\end{figure*}

Figure~\ref{fig:qualitative_heatmaps} presents representative examples of correct, partially correct, and incorrect predictions. Cases 1--4 show mostly reliable categorical attribute extraction. In Cases 1 and 3, the model correctly predicted anatomical location, margin, and attenuation, while the predicted diameter differed slightly from the reference measurement. Case 2 and Case 4 showed complete agreement with the reference attributes. These examples suggest that the model can often extract key categorical nodule features when the relevant nodule region is visually clear in the heatmap.

Case 5 shows a partial error pattern. The model correctly identified the anatomical location and margin, but it overestimated the diameter and misclassified the attenuation type. This suggests that the model may focus on a broader visual region or surrounding opacity when estimating size and density. Case 6 shows a more severe failure pattern, where the model output did not match the reference labels for anatomical location, margin, or attenuation, and the diameter estimate was also substantially different. This case indicates that model errors can increase when the heatmap attention is not well aligned with the true nodule region or when the visual appearance of the nodule is ambiguous.

\subsection*{Comparison with Baseline Models}

To contextualize the performance of the proposed fine-tuned Florence-2 model, a comparative analysis was conducted against multiple reference baselines, including a pre-trained GPT-4 model, a domain-specific fine-tuned GPT-4 model \cite{GPT4}, and available human reference values from prior radiological studies. The GPT-4 baselines were evaluated using the same test set and evaluation protocol to assess the effect of task-specific fine-tuning and to compare model behaviour across different vision-language architectures. Prior work on pulmonary nodule measurement has shown that manual diameter assessment is subject to interreader variability, and that reported overall variability in manual diameter measurement used here as the human reference for diameter estimation \cite{Han2018}. Similarly, prior observer studies have demonstrated that attenuation-related classification, such as differentiating solid from subsolid nodules, shows variability even among experienced thoracic radiologists; therefore, the reported range of radiologist performance was used to contextualize attenuation classification \cite{Ridge2016}. This comparative framework allows the proposed model to be evaluated not only against computational baselines but also against reported human variability in clinically relevant CT nodule assessment tasks. The overall comparison is summarized in Table~\ref{tab:comparative_performance}.

As shown in Table~\ref{tab:comparative_performance}, the proposed fine-tuned Florence-2 model achieved the best overall performance across all evaluated radiological attributes. For location classification, the proposed model reached 77.18\% accuracy, which was substantially higher than both the pre-trained GPT-4 baseline and the fine-tuned GPT-4 model. This indicates that the proposed model was more effective in capturing spatial information from the CT slices and map it to the correct anatomical location.

\begin{table}[htbp]
    \centering
    \caption{Performance comparison of the proposed fine-tuned Florence-2 model with GPT-4 baselines and available human reference values available in the literature for pulmonary nodule characterization tasks. For diameter, the reported value represents inter-reader variability in manual diameter measurement, whereas for attenuation it represents the reported range of radiologist performance for solid versus subsolid nodule classification.}
    \label{tab:comparative_performance}
    \begin{tabular}{|l|c|c|c|c|}
    \hline
    Attribute &
    Pre-trained GPT-4 &
    Fine-tuned GPT-4 &
    Human Reference &
    Proposed Model \\
    \hline
    Location (Accuracy) &
    36.31\% &
    43.68\% &
    -- &
    77.18\% \\
    \hline
    Diameter (MAE) &
    4.74 mm &
    3.83 mm &
    3.2 mm &
    2.58 mm \\
    \hline
    Margin (Accuracy) &
    38.42\% &
    39.47\% &
    -- &
    67.96\% \\
    \hline
    Attenuation (Accuracy) &
    37.36\% &
    63.68\% &
    58--77\% &
    79.13\% \\
    \hline
    \end{tabular}
\end{table}

For diameter estimation, the proposed model achieved the lowest MAE of 2.58 mm, compared with 4.74 mm for pre-trained GPT-4 and 3.83 mm for fine-tuned GPT-4. The proposed model also performed below the reported human interreader variability of 3.2 mm. Although the human baseline represents measurement variability rather than model prediction error, this comparison shows that the proposed model produced diameter estimates within a clinically relevant range of manual measurement variation.

For margin characterization, the proposed model achieved 67.96\% accuracy, while the pre-trained and fine-tuned GPT-4 models achieved 38.42\% and 39.47\%, respectively. The limited improvement after GPT-4 fine-tuning suggests that margin classification remains difficult for GPT-4-based models. In contrast, the stronger performance of the proposed model indicates improved recognition of boundary-related visual features.

For attenuation classification, the proposed model achieved 79.13\% accuracy, exceeding both GPT-4 baselines and the reported human baseline range of 58 -- 77\%. This result is notable because attenuation classification is visually challenging, particularly when solid and subsolid appearances overlap in a single 2D CT slice. Overall, the results show that the proposed fine-tuned Florence-2 model provides stronger and more consistent performance than the GPT-4 baselines, while also reaching a level comparable to available human-reference values for diameter and attenuation assessment.

\subsection*{Slice Selection Sensitivity Analysis}

\begin{figure}[htbp]
    \centering
    \includegraphics[width=\linewidth]{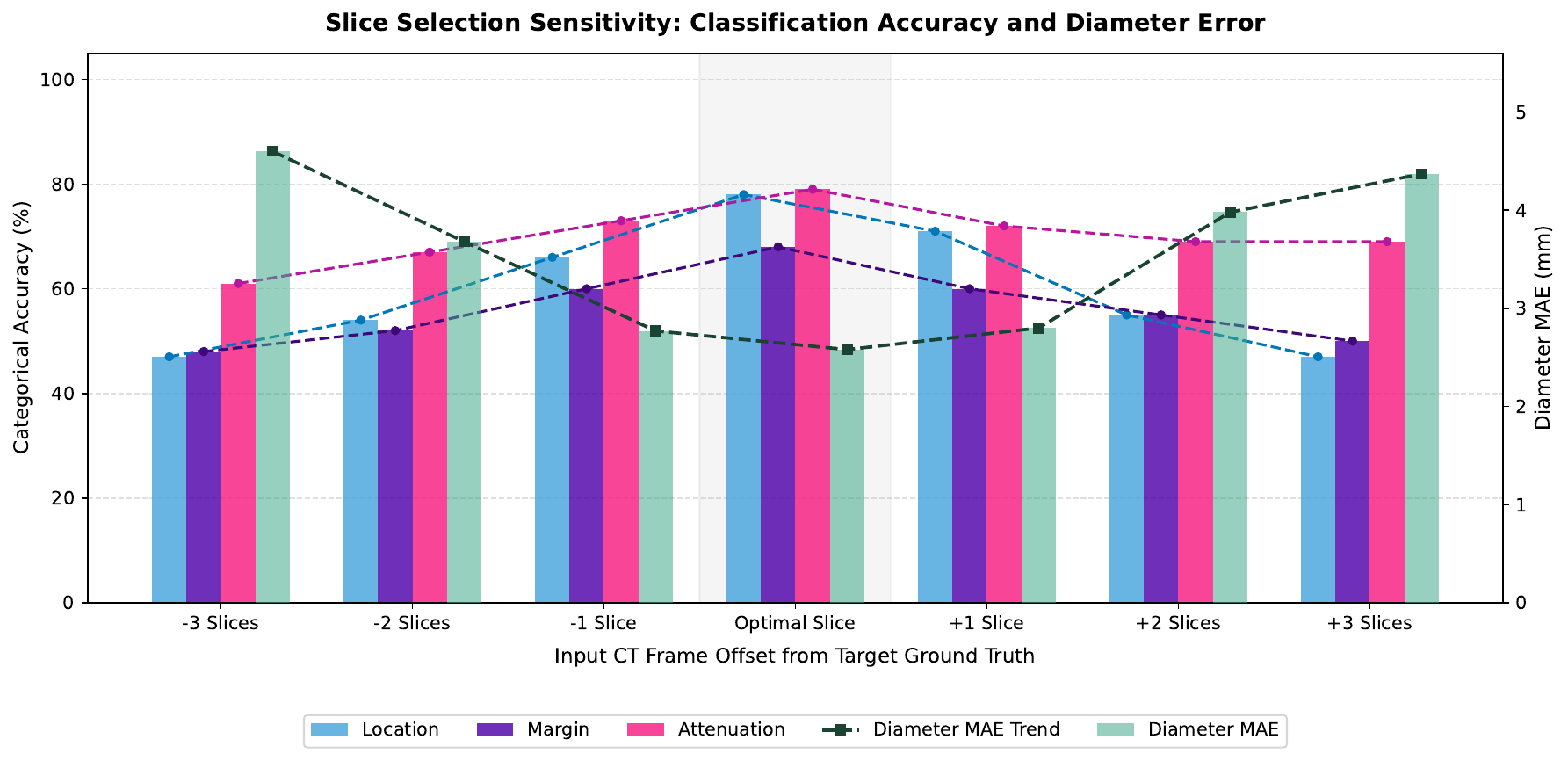}
    \caption{Performance of the finetuned Florence-2 model across different input slice offsets relative to the expert annotated reference slice.}
    \label{fig:sensitivity}
\end{figure}

Slice-selection sensitivity was evaluated by shifting the input up to three slices in both directions from the expert-annotated reference slice and measuring the resulting changes in model performance. Figure~\ref{fig:sensitivity} shows that all four prediction tasks achieved their best performance when the expert-annotated slice was used as input. As the slice moved away from this reference position, model performance generally declined, with the poorest results occurring at an offset of three slices. Anatomical location showed the greatest sensitivity, with accuracy decreasing from 78\% to 47\% at both $-3$ and $+3$ slices. Margin classification accuracy decreased from 68\% to 48\% at $-3$ slices, while attenuation was comparatively more robust, declining from 79\% to 61\% at the same offset. Diameter estimation also became less accurate, with the MAE increasing from 2.58 mm on the reference slice to 4.60 mm at $-3$ slices. These findings indicate that the model performs most reliably when the input contains the most representative nodule appearance, while larger slice offsets reduce the visual information required for accurate characterization.

\subsection*{External Generalization Assessment}

External generalization was examined using an independent veterinary CT dataset collected from Ontario Veterinary College. The dataset contained 16 CT slices, each with one pulmonary nodule and annotations for anatomical location, diameter, margin characteristics, and attenuation type. The same prompt structure and output categories used for the NLST dataset were applied, producing 64 external image-question-answer triplets. These images were not used during training or validation, and no additional finetuning was performed. Therefore, the experiment served as a preliminary domain-shift assessment of the Stage 1 Florence-2 model rather than a full external clinical validation.

The model achieved 43.75\% accuracy for anatomical location, 87.5\% accuracy for both margin characteristics and attenuation type, and an MAE of 3.75 mm for diameter estimation. The lower location accuracy likely reflects species-related differences in lung anatomy and the use of human lobar output categories for veterinary images. Margin and attenuation appeared more transferable across the two imaging domains, although their high accuracies may have been influenced by the larger proportion of smooth-margin and soft-tissue cases in the small external dataset. Overall, the results provide preliminary evidence that some learned nodule features can transfer to an independent CT source, while also showing that broader generalization claims require larger and more diverse external datasets.

\subsection*{Stage 2 Clinical Interpretation Performance}

Stage 2 was evaluated by expert human assessment across 88 generated outputs, including nodule description, follow-up recommendation, and longitudinal temporal analysis. The evaluation assessed accuracy, clinical relevance, completeness, and clinical safety.

\subsubsection*{Overall Human Evaluation}

\begin{figure}[ht]
    \centering
    \includegraphics[width=0.8\linewidth]{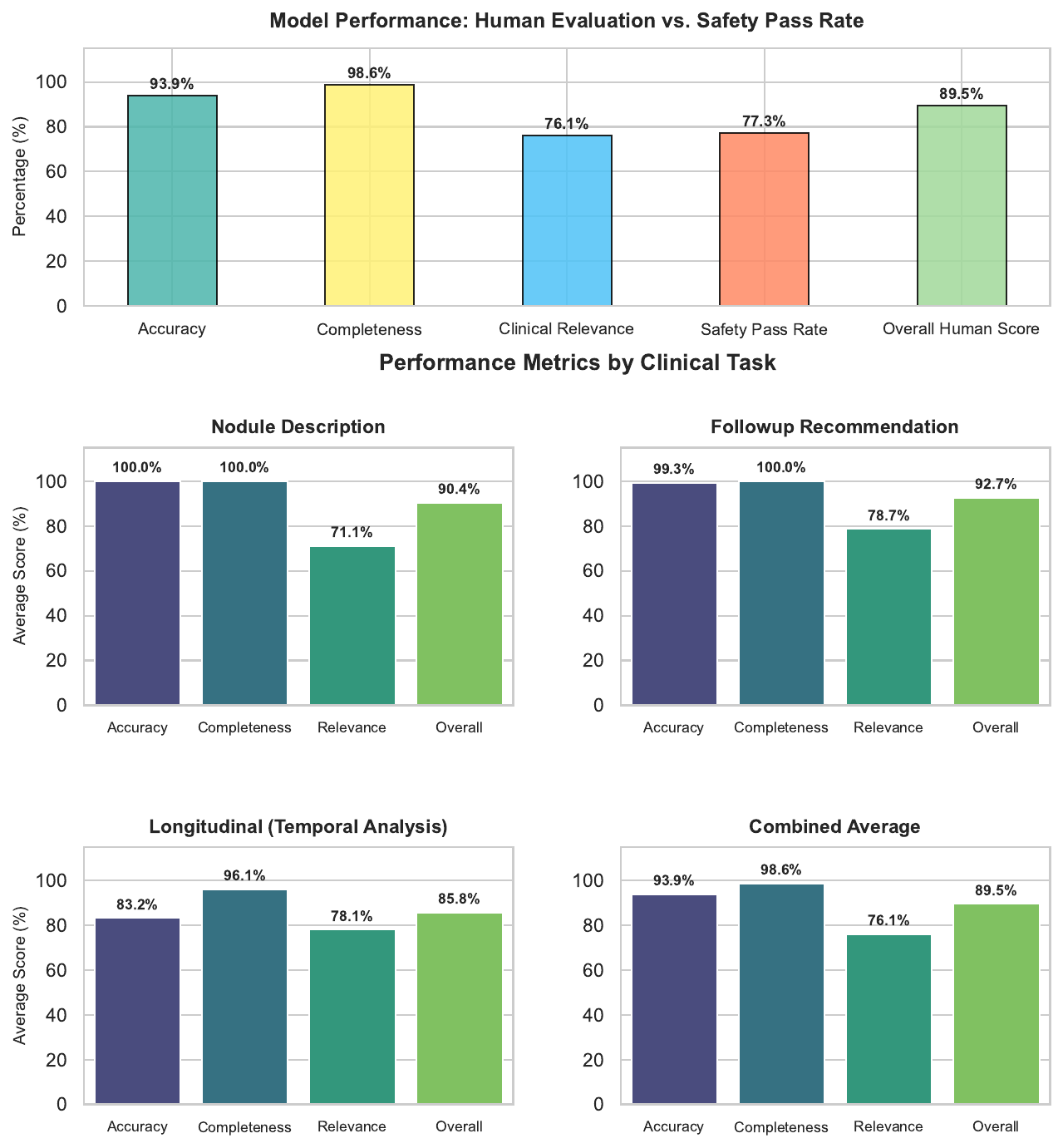}
    \caption{Human evaluation results for Stage 2 outputs across nodule description, follow-up recommendation, longitudinal temporal analysis, and combined average performance.}
    \label{fig:stage2_task}
\end{figure}

The Stage 2 model showed strong overall performance in the expert evaluation. As shown in Figure~\ref{fig:stage2_task}, the model achieved 93.8\% accuracy, 98.6\% completeness, 76.2\% clinical relevance, and an overall human score of 89.5\%. These results indicate that the model was generally able to generate clinically meaningful outputs that were consistent with the structured input attributes. The highest score was observed for completeness, suggesting that the model included most of the required clinical information in its responses. Accuracy was also high, showing that most outputs correctly reflected the given nodule characteristics.

Clinical relevance received a comparatively lower score than accuracy and completeness. This suggests that, although the generated outputs were usually correct and complete, some responses required stronger clinical prioritization or more precise recommendation language. The safety pass rate was 77.3\%, which indicates that most outputs were clinically acceptable, but a subset still required expert review before use in a clinical setting. Therefore, Stage 2 should be interpreted as an assistive clinical reasoning component rather than an autonomous decision-making system.

\subsubsection*{Performance Across Clinical Interpretation Tasks}

Task-wise evaluation showed that model performance varied across the three clinical interpretation tasks. As shown in Figure~\ref{fig:stage2_task}, nodule description achieved 100.0\% accuracy, 100.0\% completeness, 71.1\% relevance, and a 90.4\% overall score. Follow-up recommendation achieved the highest overall task score, with 99.3\% accuracy, 100.0\% completeness, 78.7\% relevance, and a 92.7\% overall score. Longitudinal temporal analysis was more challenging, with 83.2\% accuracy, 96.1\% completeness, 78.1\% relevance, and an 85.8\% overall score.

These results show that nodule description was the most stable task in terms of accuracy and completeness. This is expected because nodule description mainly requires direct conversion of structured attributes, such as location, diameter, margin, and attenuation, into descriptive clinical text. Follow-up recommendation also performed well, which suggests that the model could use the structured nodule attributes to generate reasonable management-oriented outputs. However, this task still requires caution because follow-up recommendation has direct clinical implications and depends on guideline interpretation.

Longitudinal temporal analysis showed the lowest accuracy and overall score among the three tasks. This task is more complex because the model must compare findings across multiple screening rounds, identify interval changes, and summarize temporal progression. Therefore, errors in this task may arise from difficulty in tracking multiple nodules, interpreting changes over time, or assigning appropriate clinical significance to evolving nodule features. Overall, the task-wise results suggest that Stage 2 performs well when the input structure is simple and direct, but performance becomes more variable when the task requires higher-level temporal reasoning.

\subsubsection*{Summary of Stage-2 Performance Evaluation}

The Stage 2 evaluation demonstrates that the Zephyr-based clinical reasoning component can generate accurate and complete outputs when it is grounded in structured radiological attributes. The model performed especially well for nodule description and follow-up recommendation, while longitudinal temporal analysis remained more difficult. The lower clinical relevance and safety pass rate also show that expert review remains necessary, especially for outputs involving follow-up planning or temporal disease assessment. These findings support the use of Stage 2 as a structured clinical interpretation aid, but not as a replacement for radiologist judgment.

\section*{Discussion}

This study developed and evaluated FZ-VLM, a two-stage Vision Language Model framework for structured pulmonary nodule characterization and clinical decision support. The main finding is that separating visual attribute extraction from language-based reasoning provides a traceable connection between CT image evidence and clinical interpretation. Florence-2 extracted multiple attributes within a unified model, while Zephyr-7B used these attributes for downstream clinical interpretation. This design differs from task specific approaches that require separate models for individual radiological attributes and from direct image-to-report systems in which the source of generated clinical statements may be difficult to verify.

Stage 1 demonstrated that a general-purpose VLM could be adapted for specialized lung CT interpretation. The model achieved 77.18\% accuracy for anatomical location and for diameter estimation achieved a mean absolute error of 2.58 mm, with a median absolute error of 1 mm and 72.3\% of predictions falling within $\pm$2 mm of the reference measurement. These findings indicate that most estimates remained close to the annotated values, although small errors may still be clinically important when the measured diameter is near a management threshold.

\begin{figure}[ht]
    \centering
    \includegraphics[width=\textwidth]{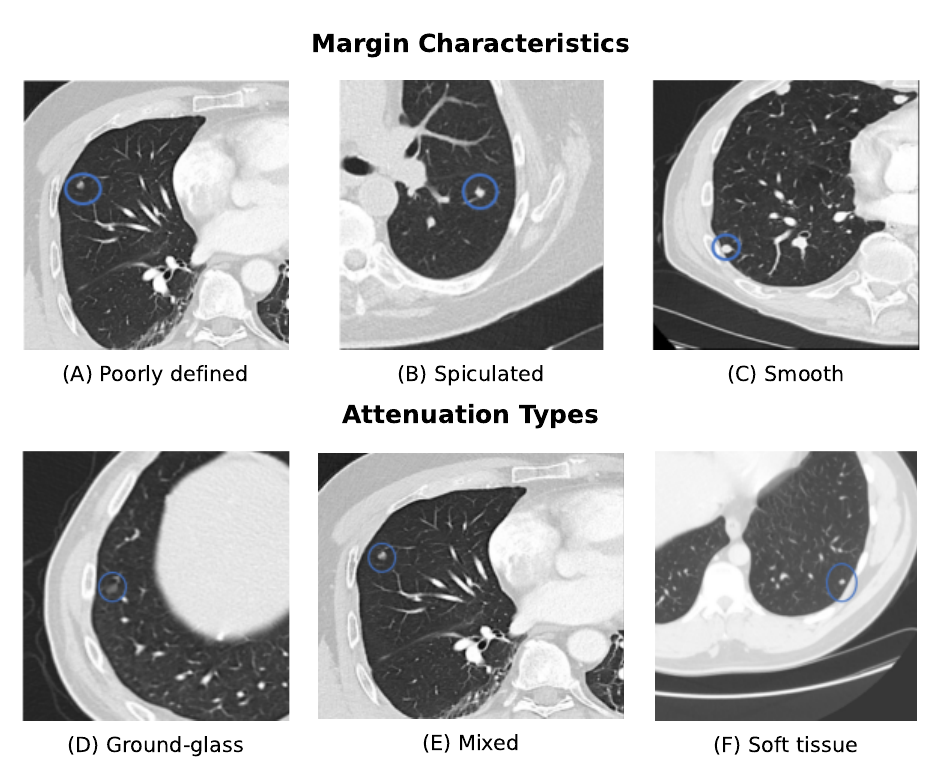}
    \caption{Representative pulmonary nodule margin characteristics and attenuation types on axial CT images. The upper row shows different margin characteristics, while the lower row shows various attenuation types. The examples illustrate the visual overlap between margin characteristics and attenuation types.}
    \label{fig:margin_attenuation}
\end{figure}

Margin and attenuation assessment depended on subtle visual differences. As illustrated in Figure~\ref{fig:margin_attenuation}, poorly defined and spiculated margins can share irregular boundary features, while mixed attenuation contains both ground-glass and soft-tissue components. These overlapping appearances reduce the distinction between classes and may also contribute to disagreement during expert assessment. The observed errors therefore likely reflect both model limitations and the inherent ambiguity of these radiological features.

The attention heatmaps provide an important interpretability strength of the proposed framework. In addition to generating radiological attribute predictions, the model highlights the image regions that contributed most strongly to its decisions. Correct predictions were generally associated with attention around the nodule or the surrounding lung region, indicating that the model used clinically relevant visual information. This feature reduces the black box nature of the model and provides experts with visual evidence that can be reviewed alongside the predicted attributes. Therefore, the heatmaps improve the transparency of Stage 1 and support the use of the framework as an assistive tool in which the expert can examine both the prediction and decision making process of the model.

The slice-sensitivity analysis identified an important practical limitation of the two-dimensional approach. Performance was highest on the expert-selected reference slice and generally decreased as the input moved farther from this position. Changes in slice position can alter the visible nodule size, boundary, attenuation pattern, and anatomical context. The results therefore support the current use of representative key slices, while also showing that automated slice selection or volumetric analysis will be needed for a more independent clinical workflow.

The baseline comparison provides strong evidence for the value of task specific adaptation. The finetuned Florence-2 model outperformed both the pretrained and finetuned GPT-4 baselines across all evaluated radiological attributes. More importantly, the proposed model also exceeded the available human reference values for attenuation classification and achieved a lower diameter error than the reported inter-reader variability. In contrast, both of the GPT-4 baselines remained below the reported human range for attenuation and showed higher diameter errors. These findings indicate that the proposed model not only improved over general purpose VLM baselines, but also reached or exceeded available human reference performance for clinically important nodule characteristics which was not achieved by the evaluated GPT baselines.

The veterinary CT experiment provided a preliminary assessment under substantial domain shift. Margin and attenuation performance remained high, whereas anatomical location accuracy decreased considerably, likely because location depends on species-specific lung anatomy and lobar organization. The findings suggest that some visual characteristics may transfer beyond the NLST data. However, the small sample size, different anatomy, and uneven class distribution prevent broader generalization and do not represent external human clinical validation.

Stage 2 results showed that structured attributes can provide an effective input for clinical language generation. Expert evaluation produced high accuracy, completeness, and overall scores, indicating that Zephyr-7B generally preserved the supplied nodule information when generating descriptions and recommendations. The high completeness score is particularly relevant because the hierarchical integration layer was designed to organize information at the nodule, study-year, and patient levels before clinical interpretation. This structured organization appears to have supported consistent generation across the three tasks.

Performance differed across the three tasks because they required different levels of reasoning. Nodule description was the most stable task because it mainly required conversion of structured attributes into clinical text. Follow-up recommendation also performed well, indicating that the model could combine multiple findings into management-oriented outputs. Longitudinal analysis was more difficult because it required tracking nodules across screening years and interpreting interval changes. Structured inputs therefore improved grounding but did not fully resolve complex temporal reasoning.

Clinical relevance and safety were lower than accuracy and completeness for different reasons. Nodule descriptions were generally accurate and complete, but because the task was mainly descriptive, some outputs added limited clinical value beyond restating the supplied attributes. Safety concerns were more evident in follow-up recommendations, where inappropriate timing could affect patient management. These findings show that expert review remains necessary, particularly for follow-up and longitudinal outputs, and that future work should improve guideline integration, risk calibration, and the clinical usefulness of generated responses.

Although the proposed FZ-VLM framework demonstrated promising performance in structured pulmonary nodule characterization and clinical interpretation, several limitations remain. First, the framework relies on expert-selected two-dimensional slices rather than complete CT volumes, which limits spatial information and creates dependence on accurate key-slice selection. Second, it operates only in a post-detection setting and assumes that the pulmonary nodule has already been identified. Third, the external assessment used a small veterinary dataset rather than an independent human cohort. Fourth, the safety pass rate of 77.3\% indicates that some Stage 2 outputs, particularly follow-up recommendations, may still be clinically inappropriate and require expert review. Finally, Zephyr-7B was used through zero-shot inference and may not consistently follow institution-specific reporting practices or management guidelines.

Overall, this study shows that FZ-VLM can link structured pulmonary nodule characterization with clinically meaningful language generation within a unified two-stage framework. The strong performance of Florence-2 and the expert-evaluated outputs from Zephyr-7B support the potential of this approach as an assistive tool for post-detection interpretation. However, the remaining limitations in safety, temporal reasoning, and external validation confirm that radiologist oversight is still essential. Future work should therefore focus on automated slice selection, three-dimensional CT analysis, validation using independent human datasets, and closer integration with established clinical guidelines.

\bibliography{Reference/reference}

\section*{Acknowledgements}

This work was supported by research funding to the AI-Driven Medical Imaging \& Diagnostics Lab at the University of Guelph, including support from an Natural Sciences and Engineering Research Council of Canada (NSERC) Discovery Grant. Data used in this study were obtained from the National Lung Screening Trial (NLST), a project supported by the U.S. National Cancer Institute. This research also used high-performance computing resources provided by the Digital Research Alliance of Canada. AI-assisted writing tools were used solely and strictly for language and grammar refinement during manuscript preparation. All scientific content, experimental design, data analysis, interpretation, and conclusions were developed and verified by the authors, who take full responsibility for the accuracy and integrity of the work.

\section*{Author contributions statement}

P.D. conducted the study design, dataset preparation, image preprocessing, model development, implementation, experiments, evaluation, analysis, and manuscript preparation. J.M. provided research guidance, supported dataset access and interpretation, and contributed to the review of the methodology and results. R.M. provided clinical and radiological guidance, evaluated the model generated clinical outputs, and reviewed representative CT cases. R.A. provided and managed the veterinary CT dataset and corresponding annotations. E.U. supervised the study, contributed to the research design and interpretation of the findings, and reviewed and revised the manuscript. All authors reviewed and approved the final manuscript.

\section*{Additional information}

\subsection*{Data availability statement}

The NLST data used in this study were obtained from the National Lung Screening Trial (NLST) through the National Cancer Institute Cancer Data Access System under project number NLST-871. The data were provided under a formal Data Transfer Agreement and cannot be publicly redistributed. Researchers may apply independently for access through the National Cancer Institute Cancer Data Access System, subject to the applicable data access requirements and institutional approvals. The source code used to develop and evaluate the FZ-VLM framework is publicly available at \url{https://github.com/PramitDutta1999/Florence-Zephyr-Vision-Language-Model-FZ-VLM-Framework}. The repository does not include restricted NLST images, patient-level information, or trained model weights.

\subsection*{Competing interests}

The authors declare no competing interests.

\end{document}